\documentclass[runningheads]{llncs}
\usepackage{graphicx}
\usepackage{amsmath}
\DeclareMathOperator*{\argmin}{arg\,min}
\usepackage{tabularray}
\usepackage{array}

\newcommand{\repeatthanks}{\textsuperscript{\thefootnote}}

\begin{document}
\title{A Continual Learning Approach for Cross-Domain White Blood Cell Classification}
\titlerunning{Continual Learning for WBC Classification}
%

\author{Ario Sadafi\inst{1,2}\thanks{equal contribution}
\and Raheleh Salehi\inst{1,3}\repeatthanks
\and Armin Gruber \inst{1,4} 
\and \\Sayedali Shetab Boushehri \inst{1,5}
\and Pascal Giehr \inst{3}
\and \\Nassir Navab \inst{2,6} 
\and Carsten Marr\inst{1}
\thanks{carsten.marr@helmholtz-munich.de}
}

\institute{Institute of AI for Health, Helmholtz Munich – German Research Center for Environmental Health, Neuherberg, Germany
\and
Computer Aided Medical Procedures (CAMP), Technical University of Munich, Germany
\and
Institute for Chemical Epigenetics Munich (ICEM), Department of Chemistry, Ludwig-Maximilian University Munich, Germany
\and
Laboratory of Leukemia Diagnostics, Department of Medicine III, University Hospital, Ludwig-Maximilians University Munich, Germany
\and
Data \& Analytics, Pharmaceutical Research and Early Development, Roche Innovation Center Munich (RICM), Penzberg, Germany
\and
Computer Aided Medical Procedures, Johns Hopkins University, USA
}

\authorrunning{Sadafi and Salehi et al.}

\maketitle       
\begin{abstract}
    Accurate classification of white blood cells in peripheral blood is essential for diagnosing hematological diseases. Due to constantly evolving clinical settings, data sources, and disease classifications, it is necessary to update machine learning classification models regularly for practical real-world use. Such models significantly benefit from sequentially learning from incoming data streams without forgetting previously acquired knowledge. However, models can suffer from catastrophic forgetting, causing a drop in performance on previous tasks when fine-tuned on new data.  Here, we propose a rehearsal-based continual learning approach for class incremental and domain incremental scenarios in white blood cell classification. To choose representative samples from previous tasks, we employ exemplar set selection based on the model's predictions. This involves selecting the most confident samples and the most challenging samples identified through uncertainty estimation of the model. We thoroughly evaluated our proposed approach on three white blood cell classification datasets that differ in color, resolution, and class composition, including scenarios where new domains or new classes are introduced to the model with every task. We also test a long class incremental experiment with both new domains and new classes. Our results demonstrate that our approach outperforms established baselines in continual learning, including existing iCaRL and EWC methods for classifying white blood cells in cross-domain environments. 

\end{abstract}

\section{Introduction}
Microscopic examination of blood cells is an essential step in laboratories for diagnosing haematologic malignancies and can be aided by computational diagnostic tools. 
So far, the developed classification methods have performed quite well. For instance, Matek et al. \cite{matek2019human} proposed an expert-level tool for recognizing blast cells in blood smears of acute myeloid leukemia patients that also works for accurately classifying cell morphology in bone marrow smears \cite{matek2021highly}. 
Several contributions are addressing the problem also at disease diagnosis level \cite{hehr2023explainable,sadafi2020attention}. Boldu et al. \cite{boldu2019automatic} have proposed an approach for diagnosing acute leukemia from blast cells in blood smear images. Sidhom et al. \cite{sidhom2021deep} and Eckardt et al. \cite{eckardt2022deep,eckardt2022deep2} are proposing methods based on deep neural networks to detect acute myeloid leukemia, acute promyelocytic leukemia and NPM1 mutation from blood smears of patients. All of these methods are highly dependent on the data source. 
Laboratories with different procedures and microscope settings produce slightly different images, and adapting models \cite{sadafi2019multiclass} to new streams of data arriving on a regular basis is a never-ending task. Although some domain adaptation methods have been proposed to increase the cross-domain reusability, for instance, the cross-domain feature extraction by Salehi et al. \cite{salehi2022unsupervised}, regular finetuning of models on new data in each center is still a necessity. 

When a model is trained on new data, catastrophic forgetting can occur in neural network models leading to loss of information it had learned from the previous data \cite{kemker2018measuring}.
Continual learning techniques are designed to tackle catastrophic forgetting and might become a necessity in clinical applications \cite{lee2020clinical}. 
They can be divided into three main categories: (i) Task-specific components where different sub-networks are designed for each task. (ii) Regularized optimization methods such as elastic weight consolidation (EWC) \cite{kirkpatrick2017overcoming} that control the update of weights to minimize forgetting. (iii) Rehearsal based approaches such as iCaRL \cite{rebuffi2017icarl} where a small memory is used to keep  examples from previous tasks to augment new datasets.

In medical image analysis, however, these continual learning techniques are less frequent.
Li et al. \cite{li2020continual} have developed a method based on continual learning in task incremental scenarios for incremental diagnosis systems. 
For chest X-ray classification, continual learning has been applied to the domain adaptation problem by Lenga et al. \cite{lenga2020continual} or on domain incremental learning by Srivastava et  al. \cite{srivastava2021continual}. 
In another work, Derakhshani et al. \cite{derakhshani2022lifelonger} benchmarked different continual learning methods on MNIST-like collections of biomedical images. Their study highlights the potential of rehearsal-based approaches in disease classification.

In this paper, we are proposing an uncertainty-aware continual learning (UACL) method, a powerful approach for both class incremental and domain incremental scenarios. We test our rehearsal-based method on three real-world datasets for white blood cell classification, demonstrating its potential. We use a distillation loss to preserve the previously learned information and provide a sampling strategy based on both representativeness of examples for each class and the uncertainty of the model. 
In order to foster reproducible research, we are providing our source code at \url{https://github.com/marrlab/UACL}

\section{Methodology}


\subsection{Problem formulation}
Let $\mathcal{D}_t = \{x_i, y_i\}$ denote the training set at step $t$ with images $x_i$ and their corresponding labels $y_i$. The classifier $f_t(.)$ is initialized with model parameters $\theta_{t-1}$ learned during the previous stage. 
To facilitate the new task (such as classifying a new class), additional parameters are introduced at every stage, which are randomly initialized.  
Our incremental learning method updates the model parameters $\theta_t$ for the new task based on the training set $\mathcal{D}_t$ and an exemplar set $X_t$. The exemplar set contains selected sample images from the previous steps.

In order to mitigate catastrophic forgetting, in UACL we suggest: (i) At every stage, the training is augmented with the exemplar set $X_t$ consisting of the sampled images from previous tasks. (ii) A distillation loss is added to preserve the information learned in the previous stage. 

\subsection{Classifier model}
Our method is architecture agnostic, and any off-the-shelf classifier can be used. We decided to follow Matek et al. \cite{matek2019human} and use a ResNeXt-50 \cite{xie2017aggregated} as the single cell classifier. ResNeXt is an extension to the ResNet \cite{he2016deep} architecture introducing the cardinality concept defined as the number of parallel paths within a building block. Each path contains a set of aggregated transformations allowing ResNeXt to capture fine-grained and diverse features.

\subsection{Training}
At every training step, the new stream of data $\mathcal{D}_t$ is combined with the examples sampled from the previous steps $X_t$. The training is performed by minimizing a total loss term $\mathcal{L}$ consisting of classification and distillation loss as follows:

\begin{equation}
    \mathcal{L}(\theta_t) = \mathcal{L}_{\mathrm{cls}}(\theta_t) + \gamma\mathcal{L}_{\mathrm{dist}}(\theta_t)
\end{equation}
where $\gamma$ is a coefficient controlling the effect of distillation on the total loss.
For classification loss, we are using categorical cross-entropy loss defined as
\begin{equation}
    \mathcal{L}_{\mathrm{cls}}(\theta_t) =  - \sum_{j \in C_t} y_i^{(j)} \mathrm{log}(f(x_i; \theta_t)^{(j)}) \qquad \forall (x_i,y_i) \in \mathcal{D}_t \cup X_t
\end{equation}
where $C_t$ is the set of all classes that the model has been exposed to since the beginning of the experiment.
The distillation loss is defined between the current state of the model with $\theta_t$ parameters and the previous state with learned $\theta_{t-1}$ parameters.
Using binary cross-entropy loss, we try to minimize the disagreement between the current model and the model before training:

\begin{multline}
    \mathcal{L}_{\mathrm{dist}}(\theta_t) =\sum_{j \in C_{t-1} }  \mathcal{S} (f(x_i; \theta_{t-1})) . \mathrm{log}( \mathcal{S}(f(x_i; \theta))) \\+ (1 - \mathcal{S} (f(x_i; \theta_{t-1}))) . \mathrm{log}(1 - \mathcal{S}(f(x_i; \theta))) \\  \forall (x_i,y_i) \in \mathcal{D}_t \cup X_t
\end{multline}
where $\mathcal{S}(x) = (1 + \mathrm{exp}(x))^{-1}$ is the sigmoid function.

\subsection{Exemplar set selection}
Selection of the exemplar set is crucial for the performance of our uncertainty aware continual learning method.  
We assume a memory budget equal to $K$ images, so for every class $m = K / |C_t|$ images can be selected. 
We use two criteria for the selection of the images:

(i) $m/2$ of the images are selected according to their distance from the class mean in the feature space such that the images that are best approximating the class mean are preserved. If $\psi(x; \theta_t)$ returns the feature vector of the model at the last convolutional layer of the model for input $x$, each image $p_i$ is selected iteratively using the following relation

\begin{equation}
    p_i =  \argmin_{x\in \mathcal{P}}\left\lVert \frac{1}{n}\sum_{x_i \in \mathcal{P}} \psi(x_i) - \frac{1}{k}[ \psi(x) + \sum_{j=1}^{k-1} \psi(p_j) ] \right\rVert
\end{equation}
where $\mathcal{P}$ is the set of all images belonging to the class and $1<k<m/2$ is the total number of selected images so far.

(ii) We use epistemic uncertainty estimation \cite{mukhoti2021deterministic} to select the remaining $m/2$ images. We estimate the uncertainty of the model on every image by measuring its predictive variance across $T$ inferences, using a dropout layer introduced between the second and third layer of the architecture. This dropout serves as a Bayesian approximation \cite{gal2016dropout}. 
For every image, the model predicts $|C_t|$ probabilities resulting into a matrix $R^{T \times |C_t| }$. Thus for every image $x_i$, the uncertainty is calculated as
\begin{equation}
     U_i = \sum_{j=1}^{|C|} \sqrt{\frac{\sum_{k=1}^{T} (R^{x_i}_{kj} - \mu^{x_i}_j)^2}{T}} \qquad \forall x_i \in \mathcal{P}
\end{equation}
where $\mu_j^{x_i} =  \frac{1}{T} \sum_{i}^{T} R^{x_i}_i$ is the vector with mean values of $R$ for each class $j$.
After this step, images are sorted based on the uncertainty scores, and the $m/2$ images with the highest uncertainty are selected.

\subsection{Exemplar set reduction} 
Since the total amount of memory is limited, in case new classes or domains are introduced, the method needs to shrink the currently preserved samples in order to accommodate the new classes. Since both lists of class mean contributors and uncertain images are sorted, this reduction is made by discarding the least important samples from the end of the list. 

\begin{figure}[t]
\centering
\includegraphics[width=\textwidth,page=1,trim=0cm 12.5cm 0.2cm 0cm,clip]{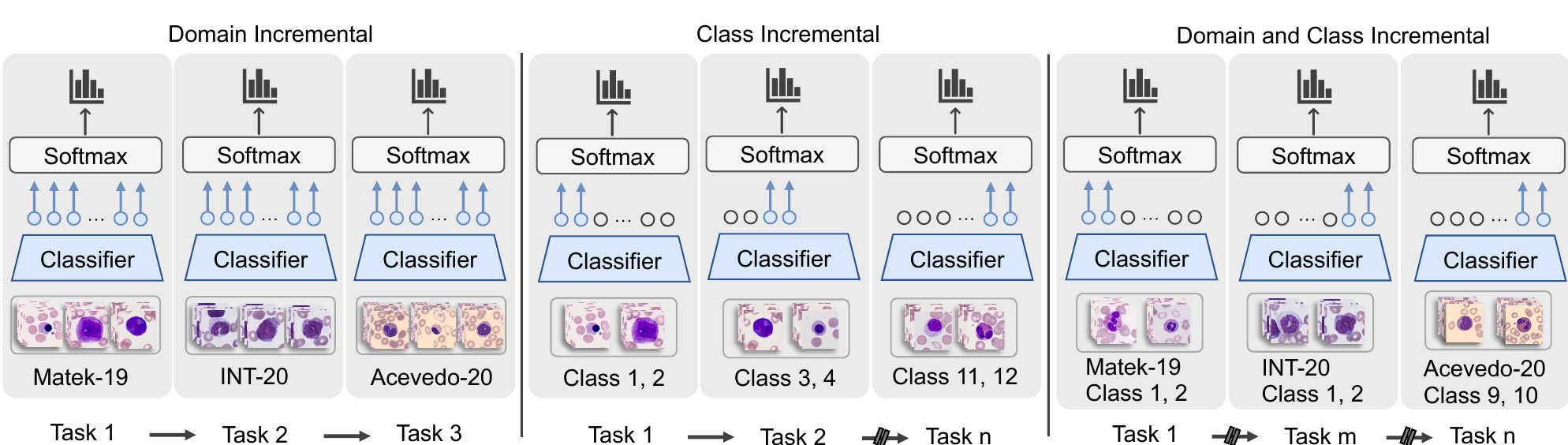}
\caption{Overview of the three different experiments we tested our proposed uncertainty aware continual learning (UACL) method: In domain incremental, in each task, the classifier is exposed to all classes of the new dataset. In class incremental, the scenario is repeated for each dataset, gradually introducing the classes to the model. Lastly, the performance of the approach is tested on a combination of domain and class incremental, where classes are gradually introduced to the model and continued with all datasets.}
\label{figoverview}
\end{figure}

\section{Evaluation}
\subsection{Datasets and continual learning experiments}
Three single white blood cell datasets from different sources are used for our continual learning experiments. 
Matek-19 \cite{matek2019single,clark2013cancer} is a single white blood cell collection
from 100 acute myeloid leukemia patients and 100 non-leukemic patients from Munich University Hospital between 2014 and 2017, and publicly available 
INT-20 is an in-house dataset with over 23,000 annotated white blood cells. Acevedo-20 is another public dataset \cite{acevedo2020dataset} with over 14,000 images, acquired at the core laboratory at the Hospital Clinic of Barcelona. 
For further details about the datasets and train test split, refer to Table \ref{tab:exp}.

\begin{table}
\centering
\caption{Statistics of the three datasets used for experiments.}
\label{tab:exp}

\begin{tblr}{
  width = \linewidth,
  colspec = {Q[160]Q[120]Q[120]Q[120]Q[110]Q[280]},
  vline{2-6} = {-}{},
  hline{2-4} = {-}{},
}
Datasets & Number of classes & Train set images & Test set images & Classes per task & Image size (pixels)\\
Matek-19 & 13 & 12,042 & 3,011 & 3,3,3,4 & $400\times400_{(29\times29 \mu m)}$\\
INT-20 & 13 & 18,466 & 4,617 & 3,3,3,4 & $288\times288_{(25\times25 \mu m)}$\\
Acevedo-20 & 10 & 11,634 & 2,909 & 3,2,2,3 & $360\times363_{(36\times36.3 \mu m)}$
\end{tblr}
\end{table}
We define three different continual learning scenarios to evaluate our method:

\textbf{- Domain incremental}: With three datasets, three different tasks are defined for training, where the model is exposed only to one of the datasets at a time and tested on the rest of the unseen datasets. The goal is to preserve the class representations while the domain changes. 

\textbf{- Class incremental}: In this scenario, four tasks are defined, each containing two to four new classes for the model to learn (see Table \ref{tab:exp}). This scenario repeats for all three datasets.

\textbf{- Domain and class incremental}
This is the longest continual learning scenario we could define with the datasets. Starting with Matek-19, the model is exposed to four tasks in a class incremental fashion. After that, the experiment continues with another 4 tasks from INT-20 followed by 4 more tasks from Acevedo-20 resulting in 12 tasks across the three datasets.

Figure \ref{figoverview} provides an overview of the scenarios, and Table \ref{tab:exp} shows statistics about the defined tasks.

\begin{table}[ht]
\caption{Average accuracy and average forgetting is reported for all methods and all experiments. Finetuning and EWC perform on par in almost all experiments, while our approach consistently outperforms the other baselines.
Abbreviations in experiment column are 
D: Domain incremental, 
C\textsubscript{M}: Class incremental Matek-19, 
C\textsubscript{I}: Class incremental INT-20, 
C\textsubscript{A}: Class incremental Acevedo-20, 
DC: Domain and class incremental.  An upper bound (UB) is also provided where models have access to all data.}
\label{tab:results}
\resizebox{\columnwidth}{!}{
\begin{tabular}{l||cc|cc|cc|cc||c}
 &
  \multicolumn{2}{|c|}{UACL} &
  \multicolumn{2}{c|}{Finetuning} &
  \multicolumn{2}{c|}{EWC} &
  \multicolumn{2}{c||}{iCaRL} &
  UB \\ \hline
 &
  \multicolumn{1}{|c|}{Acc $\uparrow$} &
  For $\downarrow$ &
  \multicolumn{1}{c|}{Acc $\uparrow$} &
  For $\downarrow$ &
  \multicolumn{1}{c|}{Acc $\uparrow$} &
  For $\downarrow$ &
  \multicolumn{1}{c|}{Acc $\uparrow$} &
  For $\downarrow$ &
  Acc$\uparrow$ \\ \hline
D &
  \multicolumn{1}{|c|}{\textbf{0.82}±0.04} &
  \textbf{0.01}±0.02 &
  \multicolumn{1}{c|}{0.57±0.28} &
  0.47±0.48 &
  \multicolumn{1}{c|}{0.58±0.28} &
  0.48±0.48 &
  \multicolumn{1}{c|}{0.80±0.09} &
  0.06±0.03 &
  0.96±0.01 \\ \hline
C\textsubscript{M} &
  \multicolumn{1}{|c|}{\textbf{0.74}±0.17} &
  \textbf{0.06}±0.10 &
  \multicolumn{1}{c|}{0.47±0.23} &
  0.28±0.39 &
  \multicolumn{1}{c|}{0.46±0.20} &
  0.24±0.26 &
  \multicolumn{1}{c|}{0.62±0.18} &
  0.08±0.12 &
  0.89±0.05 \\ \hline
C\textsubscript{I} &
  \multicolumn{1}{|c|}{\textbf{0.85}±0.11} &
  \textbf{0.08}±0.03 &
  \multicolumn{1}{c|}{0.51±0.29} &
  0.33±0.47 &
  \multicolumn{1}{c|}{0.65±0.28} &
  0.33±0.13 &
  \multicolumn{1}{c|}{0.81±0.15} &
  0.11±0.08 &
  0.97±0.02 \\ \hline
C\textsubscript{A} &
  \multicolumn{1}{|c|}{\textbf{0.83}±0.11} &
  0.05±0.04 &
  \multicolumn{1}{c|}{0.49±0.30} &
  0.33±0.47 &
  \multicolumn{1}{c|}{0.61±0.26} &
  0.16±0.05 &
  \multicolumn{1}{c|}{0.81±0.13} &
  \textbf{0.04}±0.01 &
  0.97±0.02 \\ \hline
DC &
  \multicolumn{1}{|c|}{\textbf{0.80}±0.06} &
  \textbf{0.01}±0.05 &
  \multicolumn{1}{c|}{0.22±0.23} &
  0.08±0.24 &
  \multicolumn{1}{c|}{0.23±0.23} &
  0.07±0.26 &
  \multicolumn{1}{c|}{0.52±0.16} &
  0.06±0.17 &
  0.75±0.16
\end{tabular}
}
\end{table}
\begin{figure}[t]
\centering
\includegraphics[width=0.90\textwidth,page=2,trim=0cm 1cm 0cm 0cm,clip]{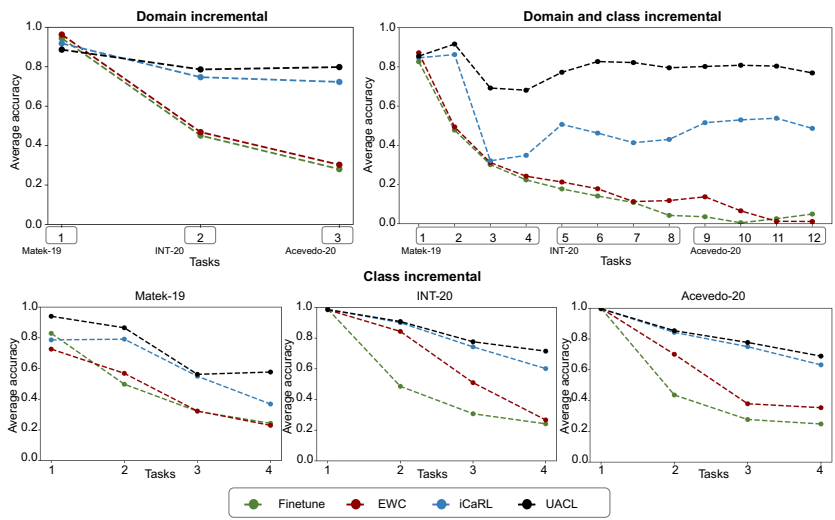}
\caption{We compare our proposed method with three different approaches as baselines for all three experiments regarding average accuracy. Our UACL method outperforms all other methods and is on par with iCaRL in domain incremental and class incremental experiments.}
\label{figeval}
\end{figure}
\begin{figure}[!ht]
\centering
\includegraphics[width=0.5\textwidth,page=3,trim=0cm 7.7cm 11cm 0cm,clip]{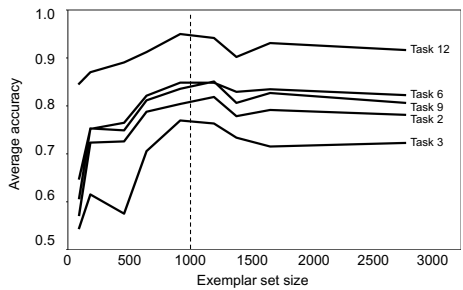}
\caption{Average accuracy over the memory of the total exemplar sets (K) in domain class incremental experiment. We conducted our experiments at K=1000.}
\label{fig:samplesize}
\end{figure}
\subsubsection{Baselines}
For comparison, we are using three different baselines: 
(i) Simple fine-tuning of the model as a naive baseline. 
(ii) Elastic weight consolidation (EWC) \cite{kirkpatrick2017overcoming} method imposes penalties on change of certain weights of the network that are important for the previously learned tasks while allowing other less important weights to be updated during training to accommodate the learning of the new task.
(iii) Incremental classifier and representation learning (iCaRL) \cite{rebuffi2017icarl}, a method combining feature extraction, nearest neighbor classification, and exemplar storage for rehearsing previous tasks.

\subsubsection{Implementation details}
We divided the datasets into stratified 75\% training set and 25\% test set. 
Models are trained for 30 epochs with stochastic gradient descent, a learning rate of $5e-4$, and a Nestrov momentum \cite{sutskever2013importance} of $0.9$. 
Models are initialized with ImageNet \cite{deng2009imagenet} weights for the training of the first task. 
For EWC experiments, the learning rate is set to $3e-4$ and lambda to 0.9 to weight EWC penalty loss. EWC penalty importance is set to $3000$.  
For iCaRL experiments, we decided to have an exemplar size $K = 1000$ and  $\gamma = 1$ for 3-dataset experiments (i.e., domain incremental and class-domain incremental experiments), while $K = 125$ and $\gamma = 5$ for class incremental experiments on single datasets. We used $10$ forward passes for MC dropout sampling.
We use average accuracy \cite{de2021continual} and average forgetting \cite{de2021continual} for the evaluation of all experiments.

\subsection{Results and Discussion}
We have conducted five experiments at class incremental and domain incremental learning (see Table \ref{tab:results}). In addition to the defined baselines (Finetuning, EWC, and iCaRL), we are also comparing our UACL method to the hypothetical scenario where the model has access to complete datasets from previous tasks, as an upper bound (UB) to the problem.  Our UACL method outperforms all other baselines in average accuracy (Acc). It outperforms all other baselines also in average forgetting (For), apart from the class incremental Acevedo-20 experiment, where iCaRL is slightly better (0.04 vs. 0.05, see Table \ref{tab:results}). 

Our UACL method is specifically good for long-term learning. According to Figure \ref{figeval}, UACL outperforms all other methods in the domain class incremental experiments due to its epistemic uncertainty-based sampling that correctly samples the most challenging examples from every class.
Interestingly, we are outperforming the upper bound in the domain class incremental experiment, which can be due to the unbalanced data (see Table \ref{tab:results}). With the UACL sampling strategy, classes are equally sampled, and uncertain cases in underrepresented classes are very informative to preserve the latent structure of these classes.

An important hyperparameter in the experiments is the exemplar set size $K$. We conducted an experiment studying the effect of $K$ on the average accuracy metric. Figure \ref{fig:samplesize} demonstrates the average accuracy for different values of $K$ for different tasks in the domain and class incremental experiment. As expected, small values of the exemplar set size are unable to capture the variability present in the classes. However, as the value of $K$ increases, the method reaches higher average accuracy and a plateau. Specifically, the experiment indicates that an exemplar set size of around $K=1000$ achieves the optimal value for capturing class variability and achieving high accuracy while using less memory.

\begin{figure}[t]
\centering
\includegraphics[width=0.90\textwidth,page=4,trim=0cm 7cm 0cm 0cm,clip]{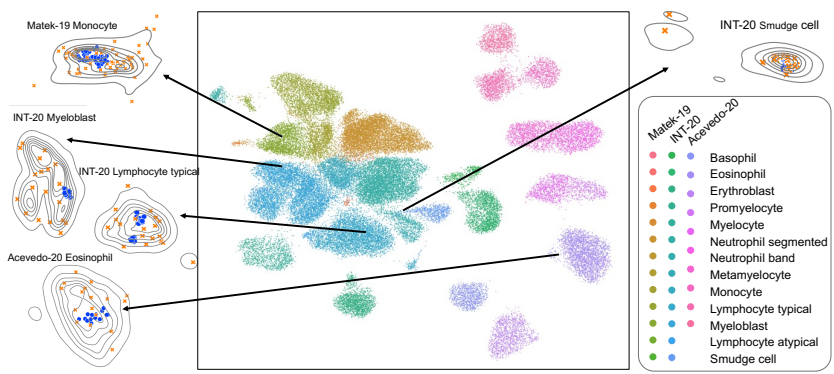}
\caption{Epistemic uncertainty-based sampling better preserves the latent distribution of the classes. This is demonstrated with the UMAP embedding of the training set with a total of 52,629 cells. Each class in each dataset has a different color. For some classes, the kernel distribution estimation plot and the selected examples by epistemic UCAL sampling (orange) and iCaRL-based sampling (blue) are displayed.
}
\label{figeumap}
\end{figure}

Additionally, we tried to qualitatively study the difference in the UACL sampling method we are suggesting. Figure \ref{figeumap} shows the UMAP of the features on the penultimate layer of the ResNext model. We have zoomed into some of the class distributions and show the difference between our sampling approach (orange) and iCaRL's class mean preservation approach (blue). Our uncertainty-based sampling leads to a better distribution of the samples within a class. We believe scattered points over the whole class distribution and closer points to class distribution borders are essentials for long-term continual learning.

\section{Conclusion}

Accurate classification of white blood cells is crucial for diagnosing hematological diseases, and continual learning is necessary to update models regularly for practical, real-world use with data coming from different domains or new classes. To address this issue, we proposed a rehearsal-based continual learning approach that employs exemplar set selection based on the model's uncertainty to choose representative samples from previous tasks. We defined three continual learning scenarios for comparison. The results indicate that our proposed method outperforms established baselines in continual learning, including iCaRL and EWC methods, for classifying white blood cells. 
Collecting larger and better curated datasets to perform chronological incremental learning where data is collected at different time points in a center, performing longer experiments to investigate the scalability of our proposed method, and investigation of other uncertainty estimation methods are some of the exciting directions for possible future works.

\section*{Acknowledgments}
C.M. has received funding from the European Research Council (ERC) under the European Union’s Horizon 2020 research and innovation programme (Grant agreement No. 866411). R.S. and P.G. are funded by DFG, SFB1309 - project number 325871075

\bibliographystyle{splncs04}
\bibliography{article}
\end{document}